\documentclass[conference]{IEEEtran}
\IEEEoverridecommandlockouts
\usepackage{cite}
\usepackage{amsmath,amssymb,amsfonts}
\usepackage{amsthm}
\usepackage{algorithmic}
\usepackage{algorithm}
\usepackage{graphicx}
\usepackage[export]{adjustbox}
\usepackage{textcomp}
\usepackage{xcolor}
\usepackage{booktabs}
\usepackage{tabularx}
\newcolumntype{R}{>{\raggedleft\arraybackslash}X}
\newcolumntype{C}{>{\centering\arraybackslash}X}
\newcolumntype{L}{>{\raggedright\arraybackslash}X}
\usepackage{float}

\definecolor{rankgold}{HTML}{FFD700}
\definecolor{ranksilver}{HTML}{708090}
\definecolor{rankbronze}{HTML}{8C5A2B}
\newcommand{\rkone}[1]{\textcolor{rankgold}{\textbf{#1}}}
\newcommand{\rktwo}[1]{\textcolor{ranksilver}{\underline{#1}}}
\newcommand{\rkthree}[1]{\textcolor{rankbronze}{\textit{#1}}}

\makeatletter
\long\def\@makecaption#1#2{%
\ifx\@captype\@IEEEtablestring%
\footnotesize\bgroup\par\centering\@IEEEtabletopskipstrut%
{\normalfont\footnotesize #1}\\{\normalfont\footnotesize #2}\par
\addvspace{0.5\baselineskip}\egroup%
\@IEEEtablecaptionsepspace
\else
\@IEEEfigurecaptionsepspace
\setbox\@tempboxa\hbox{\normalfont\footnotesize {#1.}\nobreakspace\nobreakspace #2}%
\ifdim \wd\@tempboxa >\hsize%
\setbox\@tempboxa\hbox{\normalfont\footnotesize {#1.}\nobreakspace\nobreakspace}%
\parbox[t]{\hsize}{\normalfont\footnotesize\noindent\unhbox\@tempboxa#2}%
\else%
\ifCLASSOPTIONconference \hbox to\hsize{\normalfont\footnotesize\hfil\box\@tempboxa\hfil}%
\else \hbox to\hsize{\normalfont\footnotesize\box\@tempboxa\hfil}%
\fi\fi\fi}
\newcommand{\linebreakand}{%
\end{@IEEEauthorhalign}%
\hfill\mbox{}\par%
\mbox{}\hfill\begin{@IEEEauthorhalign}%
}
\makeatother

\begin{document}

\title{Nominate-then-Adjudicate: LLM-Assisted One-Pass and
Per-Instance MIP Solver Tuning}

\author{
\IEEEauthorblockN{Yifan Zhao}
\IEEEauthorblockA{\textit{Alibaba Group} \\
Hangzhou, China\\
jinxiong.zyf@alibaba-inc.com}
\and
\IEEEauthorblockN{Qianyang Zhao}
\IEEEauthorblockA{\textit{Alibaba Group} \\
Hangzhou, China\\
zhaoqianyang.zqy@alibaba-inc.com}
\linebreakand
\IEEEauthorblockN{Xinyu Guan}
\IEEEauthorblockA{\textit{Alibaba Group} \\
Hangzhou, China\\
guanhan.gxy@alibaba-inc.com}
\and
\IEEEauthorblockN{Kai Wei}
\IEEEauthorblockA{\textit{Alibaba Group} \\
Hangzhou, China\\
weikai.weikai@taobao.com}
\and
\IEEEauthorblockN{Yuming Deng}
\IEEEauthorblockA{\textit{Alibaba Group} \\
Hangzhou, China\\
yuming.dym@taobao.com}
}

\maketitle

\begin{abstract}
Mixed-integer programming solvers expose hundreds or even thousands of parameters whose
best setting varies per instance, yet per-instance search is prohibitively expensive. We
frame this as configuration for heterogeneous Big Data optimization workloads and propose
a nominate-then-adjudicate pipeline built around two design choices. First, a
fixed portfolio is sampled once and reused across instances, removing per-instance
configuration search at inference; on the evaluated families, the resulting shortlist
retains substantial oracle headroom, making selection within it the empirical focus.
Second, the fixed $20$-feature representation can map
structurally different instances to the same vector, preventing a feature-only selector
from distinguishing them even when they prefer different configurations. The foundation
model receives no task-specific weight updates, and selection is one-pass: one model call
and no solver calls at inference. The study represents structural Variety through distinct
formulation families and decision Value through solver-performance outcomes; the one-call,
zero-solve protocol specifies the decision interface and leaves end-to-end latency unmeasured.
Among comparisons exceeding a $20\%$ practical-difference threshold, the complete
adjudicator records $538$ wins and $185$ losses against the top nominee, while its win
rate against the default rises from $84.9\%$ to $90.2\%$. Against per-instance configuration baselines
under a matched one-pass protocol it leads the cluster-level
methods and the best-performing evaluated per-instance baseline, ISAC+.
\end{abstract}


\begin{figure*}[t]
\centering
\includegraphics[width=\textwidth]{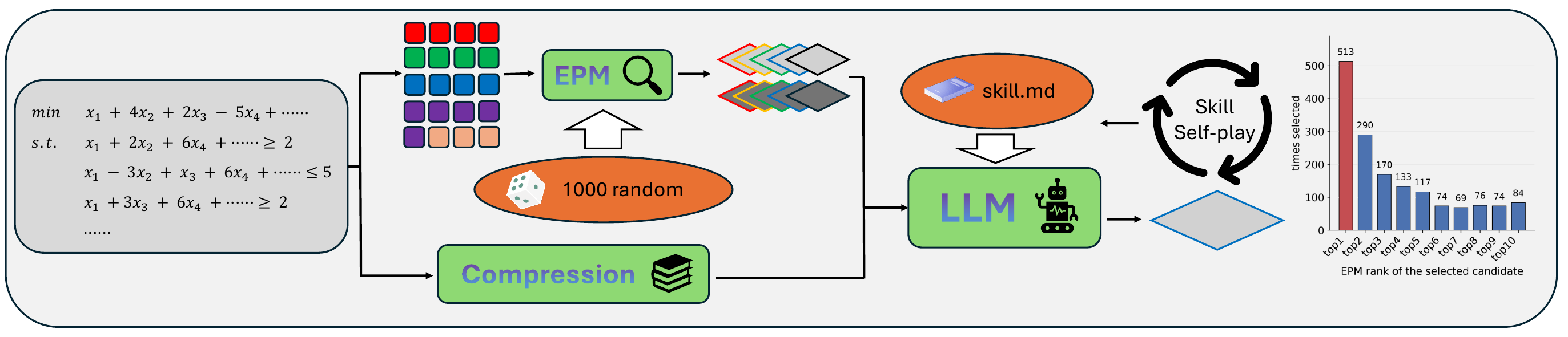}
\caption{The nominate-then-adjudicate pipeline. The instance is read twice: one path
abstracts it into $20$ features that, with each candidate's $30$ parameters, are all the
EPM sees; the other passes a compressed constraint matrix to the adjudicator. Portfolio,
EPM and skill memory are built offline, so inference is one model call and no solve.}
\label{fig:framework}
\end{figure*}

\section{Introduction}

Branch-and-cut solvers such as SCIP~\cite{bestuzheva2023scip} expose thousands of
algorithmic parameters, and no single configuration is best across instances. Such solvers
sit beneath large-scale combinatorial decision-making. Routing, assignment and
supply-chain optimization are routinely formulated as MIPs, so a per-instance
configuration decision that adds no solver calls is of practical interest.

The Big Data relevance of this setting centers on heterogeneous optimization workloads.
We operationalize Variety through $16$ structurally distinct
generated MIP families and two heterogeneous MIPLIB transfer cases, and Value through score
and solver-time improvements. Velocity enters as an interface requirement: the configuration
decision uses one model call and no online solver calls. End-to-end latency remains unmeasured.

Existing methods can be placed on three axes: the \emph{granularity} of the decision (a whole
distribution, a cluster, or a single instance), the number of \emph{solver calls} it
costs at inference, and whether a \emph{selector must be fitted}. Together, these axes
highlight an under-studied design point. Classical configurators return a distribution-wide
configuration. Per-instance selectors instead fit a mapping from features to
configurations. Methods that read more than static features pay for it with solver calls at
decision time~\cite{hutter2011smac,lopez2016irace,ansotegui2009gga,kadioglu2010isac,ansotegui2014isacplus,xu2011hydramip,song2023instance,xu2008satzilla,gasse2019exact}.
Recent language-model approaches to
configuration~\cite{lawless2025llmsep,kochnev2025optuna,wu2024asllm,liu2024llambo} also
miss at least one axis. Section~\ref{sec:related} places each precisely.

Against this landscape, we combine natural-language processing, operations research and
machine learning in one pipeline: a learned performance model ranks a fixed solver
portfolio, and a foundation model adjudicates the shortlist in \emph{one model call, with
no solver call at inference and no foundation-model weight update}. To our knowledge, this
is the first reported per-instance MIP configuration system to combine those operational
properties. Offline adaptation remains: the nominator is a small tree
ensemble trained on measured solves, and the adjudicator consults a textual memory refined
from selection outcomes. The foundation model's weights remain fixed. A representation-level observation further motivates the
dual-channel design (Section~\ref{sec:principle}): the final decision cannot always be
recovered from the aggregate features used by the nominator, because structurally
different instances can share the same feature vector.

\emph{Why a Nominate-then-Adjudicate system.}
A feature-only EPM is limited to the structure retained by its fixed aggregate feature
map and cannot recover distinctions that the map discards. Two instances alike in those
features can therefore prefer different configurations; yet the
configuration pool is far too large to present exhaustively in a compact one-pass prompt.
We therefore factorize the problem into recall
and re-ranking: a \textbf{nomination} stage scores a fixed portfolio with an empirical
performance model and forwards the top $k$; a
\textbf{adjudication} stage hands those $k$, a compressed textual rendering of the
constraint matrix and a textual skill memory to a language model, which returns one index.
The instance is thus read twice
through different channels (Fig.~\ref{fig:framework}): the features feed the nominator, and both feed the adjudicator.

\emph{Why use a random portfolio?}
Bergstra and Bengio~\cite{bergstra2012random} showed that random search can explore
important coordinates more efficiently than grid search when only a subset of dimensions
matters. This motivates a simple reusable candidate set, with no portfolio-coverage
guarantee. We construct the fixed portfolio once by random sampling and reuse it for every
instance, so the online nominator only has to rank this fixed set.

\emph{Why the second stage reads the matrix.}
Selection can fail when the fixed aggregate feature extractor maps structurally different
instances to the same vector. A deterministic selector that observes only that vector
must return the same choice for both, even when their preferred portfolio configurations
differ (Section~\ref{sec:principle}). Independent-set and vertex-cover instances built on
one graph realize such a collision under our $20$-feature representation. The adjudicator
can distinguish them because it also reads a channel those features discard: the
compressed textual matrix rendering. It emits
no parameter values; it makes a bounded $k$-way choice over measured candidates.

\emph{Contributions.}
(i) A nominate-then-adjudicate architecture that separates candidate recall from
structure-aware re-ranking and constrains the foundation model to a bounded $k$-way
decision over measured configurations, preventing unconstrained parameter generation,
with one model call and no solver call at inference.
(ii) A dual-channel selector in which aggregate features support efficient nomination,
while a compressed matrix representation and textual experience memory provide
complementary structural information for adjudication. The IS/VC construction gives a
concrete collision under the chosen $20$-feature representation.
(iii) An empirical evaluation across $16$ generated MIP families, a matched one-pass
baseline study, and two feature-range-compatible MIPLIB transfer cases. Under the evaluated
protocol, the complete adjudicator bundle improves over the feature-only top nominee, and
the full system leads the aggregate metrics in the matched baseline study. These experiments
evaluate Variety as cross-family structural heterogeneity and Value as solver performance;
the online protocol additionally enforces the stated one-call, zero-solve decision constraint.

\section{Related Work}
\label{sec:related}

\paragraph{Granularity}
Classical configurators tune a single configuration for a whole instance distribution:
SMAC~\cite{hutter2011smac,lindauer2022smac3} fits a surrogate over the parameter space,
irace~\cite{lopez2016irace} races sampled candidates, and GGA~\cite{ansotegui2009gga}
evolves a population (surveyed in~\cite{schede2022survey,hutter2014algorithm}).
Per-instance methods refine the granularity: ISAC~\cite{kadioglu2010isac} clusters
instances and tunes one configuration per cluster, a piecewise-constant map that learned
embeddings later smooth~\cite{song2023instance}, while ISAC+~\cite{ansotegui2014isacplus}
and Hydra-MIP~\cite{xu2011hydramip} pool such configurations and select among them at
instance granularity (per-instance selection is surveyed
by~\cite{kerschke2019survey}). MIP-specific configuration~\cite{hutter2010mip} and learned
SCIP separator scoring~\cite{li2023separators} are the closest MIP instantiations.
Assembling a fixed portfolio offline and selecting one member per instance from static
features is thus a classical skeleton, and we claim no novelty for it.

\paragraph{Solver calls at inference}
Methods that read more than static features pay for it with solver work at decision time:
SATzilla~\cite{xu2008satzilla} selects among solvers using runtime models whose features
include solver probes, Hydra~\cite{xu2010hydra} closes the loop between configuring and
selecting, and a broader line of in-solver ML replaces or drives a solver component with
a network invoked during search, including learning-to-branch~\cite{gasse2019exact}, neural MIP
solvers~\cite{nair2020solving} and learned heuristic scheduling in
branch-and-bound~\cite{chmiela2021learning}. The same cost recurs among language-model methods:
Lawless et al.~\cite{lawless2025llmsep} propose separator settings for a problem class that
the solver must then run to compare, and LLAMBO~\cite{liu2024llambo} places the model
inside a Bayesian-optimization loop that evaluates candidates repeatedly.

\paragraph{Whether a selector is fitted}
The per-instance selectors above are all fitted on measured runtimes, so what they can
learn is bounded by what the feature extractor kept. Language-model approaches either adapt
weights, as Kochnev et al.~\cite{kochnev2025optuna} do by fine-tuning a model to emit
hyperparameters, or feed a conventionally trained selector, as when Wu
et al.~\cite{wu2024asllm} turn model embeddings into its features. We instead keep
accumulated experience as text tied to measured selection outcomes, relating to
self-reflective and experience-accumulating agents~\cite{shinn2023reflexion,zhao2024expel}
and to prompt-level optimization~\cite{agrawal2025gepa}, so nothing in the model's weights
is fitted. The EPM and textual memory remain task-specific offline components; the
foundation-model weights remain fixed. Against this taxonomy, our contribution lies in
the per-instance, solver-free-at-inference adjudication design and in the structural and
textual information made available to the selector.

\section{Preliminaries}
\label{sec:prelim}

An instance $\mathcal{I}$ is a MIP model. A configuration
$\theta\in\Theta=\prod_{p=1}^{30}\mathcal{C}_p$ assigns each of $30$ SCIP parameters
a value from a finite candidate list $\mathcal{C}_p$; $|\Theta|\approx 7.9\times
10^{18}$. Let $t(\mathcal{I},\theta)$ be wall-clock solve time under a fixed time
limit and $\mathrm{sp}(\mathcal{I},\theta)=t(\mathcal{I},\theta_{\mathrm{def}})/
t(\mathcal{I},\theta)$ the speedup over the default.

We measure a configuration against the default by a single \textbf{score}
$s(\mathcal{I},\theta)\in[-1,1]$, defined by a lexicographic order. Feasibility is
decided first: a configuration finding no solution where the default does scores $-1$,
and conversely $+1$; if neither finds one the instance is discarded. Proven optimality
comes next, so a configuration that exhausts the time limit scores $-1$ against one that
closes the instance, and timeouts are scored rather than dropped. When both prove
optimality, which is the common case, the score is
\begin{equation}
\label{eq:score}
s(\mathcal{I},\theta)\;=\;
\frac{t(\mathcal{I},\theta_{\mathrm{def}})-t(\mathcal{I},\theta)}
     {t(\mathcal{I},\theta_{\mathrm{def}})+t(\mathcal{I},\theta)}
\;=\;\tanh\!\Big(\tfrac{1}{2}\log \mathrm{sp}(\mathcal{I},\theta)\Big),
\end{equation}
a strictly monotone bounded reparameterisation of $\log\mathrm{sp}$, the quantity
underlying geometric-mean runtime comparison; when neither side proves optimality the
same normalised difference is applied to the feasible objectives. The transformation is
zero at parity, symmetric between equal speedups and slowdowns, and depends on the
runtimes only through their ratio, so machine speed cancels. Boundedness is why we
average it: speedups are heavy-tailed, so a mean speedup is decided by a few instances,
whereas $\log\mathrm{sp}$ is unbounded below and one timeout dominates. The price is
saturation in the tails, so the score reflects how \emph{often} a better configuration is
chosen more than by how much, and we report the shifted geometric mean (SGM) of speedup
beside it as the physically interpretable figure.

\section{Method}
\label{sec:method}
\label{sec:principle}

The pipeline reads each instance twice through different channels and decides in one
pass. Algorithm~\ref{alg:pipeline} is the whole of inference: extract features, score a
fixed numbered portfolio with an empirical performance model (EPM), and make a single
language-model call that returns an index into the top $k$. No solver is run to decide.
All solver-intensive work happens offline before deployment: building the portfolio,
fitting the EPM and revising the skill memory.

\begin{algorithm}[H]
\caption{Nominate-then-Adjudicate Inference}
\label{alg:pipeline}
\begin{algorithmic}[1]
\REQUIRE Instance $\mathcal{I}$; EPM $F$; skill memory $\mathcal{S}$; portfolio $\Pi=\{\theta_0,\dots,\theta_{999}\}$
\STATE $\mathbf{x} \gets \textsc{ExtractFeatures}(\mathcal{I})$ \COMMENT{$20$-dim, Table~\ref{tab:features}}
\STATE $\hat r_j \gets F([\mathbf{x}\,\|\,\mathrm{enc}(\theta_j)])$ for all $\theta_j\in\Pi$ \COMMENT{no solves}
\STATE $\mathcal{N} = \{\theta_{n_1}, \theta_{n_2}, \cdots, \theta_{n_k}\}\gets$ top-$k$ of $\Pi$ by $\hat r_j$ \COMMENT{identifiers retained; $k=10$}
\STATE $i^\star \gets \textsc{LLMSelect}(\psi(\mathcal{I}), \mathbf{x}, \mathcal{N}, \mathcal{S})$ \COMMENT{$i^\star\in\{1..k\}$, default $1$}
\RETURN $\theta_{n_{i^\star}}$
\end{algorithmic}
\end{algorithm}

\subsection{Step 1: Static Features}
From the parsed model we extract $\mathbf{x}\in\mathbb{R}^{20}$, listed in
Table~\ref{tab:features}. All are cheap, solver-agnostic and solve-independent, and all
are permutation-invariant aggregates of the coefficient matrix. This property matters
in Step~3. The set is the \emph{static} branch of an established feature taxonomy for MIP
parameter tuning. We deliberately drop that taxonomy's root-relaxation branch, including
root solve time, reduced-cost and dual-value distributions, because obtaining those features
requires actually running the solver, which is precisely the cost our one-pass setting
refuses to pay at inference.

\begin{table}[!t]
\centering
\caption{The $20$ static features, by block.}
\label{tab:features}
\footnotesize
\begin{tabularx}{\columnwidth}{@{}l L@{}}
\toprule
block & features \\
\midrule
size (4) & row count, column count, nonzeros, density \\
variables (4) & binary / general-integer / continuous ratio, unbounded count \\
constraints (4) & equality ratio, inequality ratio, set-type count, special-constraint count \\
coefficients (5) & mean nonzeros per row, per column, coefficient CV, objective density, RHS CV \\
topology (3) & connected components, largest-component ratio, isolated variables \\
\bottomrule
\end{tabularx}
\end{table}

\subsection{Step 2: A Fixed Random Portfolio}
The parameters we tune are listed in Table~\ref{tab:params}: $30$ SCIP parameters
spanning presolving, separation, cut selection, primal heuristics, branching and
conflict analysis, each with a finite candidate list, giving
$|\Theta|\approx 7.9\times 10^{18}$. They were distilled by consensus from SCIP's full
parameter listing of over ten thousand entries. Several independent language models were
each asked which parameters most affect MIP solving performance, and we kept only the
intersection every model flagged, then assigned each survivor a finite candidate list.
This common experimental setting is shared by \emph{every} arm we report, baselines included.

Random sampling provides a simple way to construct a reusable candidate set without an
instance-specific configurator at inference. Motivated by random search in spaces where a
subset of coordinates drives performance~\cite{bergstra2012random}, we draw a pool of
$1000$ configurations once, uniformly from the candidate lists, and number them
$\Pi=\{\theta_0,\dots,\theta_{999}\}$. This sampled pool is the empirical object evaluated
below; random construction provides no near-optimal coverage guarantee for every
instance. Building the pool once amortizes candidate construction across instances, and
the fixed numbering lets rules and accumulated experience remain attached to portfolio
\emph{configurations} across batches. The pool is built from a batch
seed-disjoint from every evaluation batch.

\begin{table}[!t]
\centering
\caption{The $30$ configured SCIP parameters and their candidate values.}
\label{tab:params}
\scriptsize
\begin{tabular}{@{}ll@{}}
\toprule
parameter & candidates \\
\midrule
\multicolumn{2}{@{}l}{\emph{presolving}}\\
\texttt{presolving/maxrounds}          & $-1,0,5,10,20$ \\
\texttt{presolving/donotaggr}          & F, T \\
\texttt{presolving/donotmultaggr}      & F, T \\
\multicolumn{2}{@{}l}{\emph{LP}}\\
\texttt{lp/pricing}                    & l, a, f, p, s, q, d \\
\texttt{lp/refactorinterval}           & $0,50,100,200$ \\
\texttt{lp/scaling}                    & $0,1,2$ \\
\multicolumn{2}{@{}l}{\emph{propagation and symmetry}}\\
\texttt{propagating/maxrounds}         & $-1,0,10,50,100$ \\
\texttt{propagating/maxroundsroot}     & $-1,100,500,1000,5000$ \\
\texttt{misc/usesymmetry}              & $0$--$7$ \\
\multicolumn{2}{@{}l}{\emph{separation}}\\
\texttt{separating/maxroundsroot}      & $-1,0,5,10,20$ \\
\texttt{separating/maxstallrounds}     & $0,1,3,5$ \\
\texttt{separating/clique/freq}        & $-1,0,5,10,20$ \\
\texttt{separating/oddcycle/freq}      & $-1,0,5,10,20$ \\
\texttt{separating/gomory/freq}        & $-1,0,5,10,20$ \\
\texttt{separating/zerohalf/maxrounds} & $-1,0,1,5,10$ \\
\texttt{separating/disjunctive/maxroundsroot} & $-1,0,10,50,100$ \\
\multicolumn{2}{@{}l}{\emph{cut selection}}\\
\texttt{cutselection/hybrid/minortho}  & $0.1,0.3,0.5,0.7,0.9$ \\
\texttt{cutselection/ensemble/} & \\
\quad\texttt{maxnonzerorootround}      & $1,2,4.5,8,20$ \\
\multicolumn{2}{@{}l}{\emph{primal heuristics}}\\
\texttt{heuristics/feaspump/freq}      & $-1,0,10,20,50$ \\
\texttt{heuristics/alns/freq}          & $-1,0,10,20,50$ \\
\texttt{heuristics/undercover/freq}    & $-1,0,10,20,50$ \\
\texttt{heuristics/dins/minfixingrate} & $0.1,0.2,0.3,0.5,0.7$ \\
\texttt{heuristics/lpface/minfixingrate} & $0.05,0.1,0.2,0.3,0.5$ \\
\texttt{heuristics/subnlp/presolveemphasis} & $0,1,2,3$ \\
\multicolumn{2}{@{}l}{\emph{branching}}\\
\texttt{branching/preferbinary}        & T, F \\
\texttt{branching/fullstrong/maxdepth} & $-1,0,5,10,20$ \\
\texttt{branching/relpscost/sbiterquot} & $0.1,0.2,0.3,0.5,0.8$ \\
\texttt{branching/relpscost/} & \\
\quad\texttt{degeneracyaware}          & $0,1,2$ \\
\multicolumn{2}{@{}l}{\emph{conflict analysis}}\\
\texttt{conflict/enable}               & T, F \\
\texttt{conflict/useinflp}             & o, c, d, b \\
\bottomrule
\end{tabular}
\end{table}

\subsection{Step 3: Nomination by an EPM}
\label{sec:nominate}
Let $\mathrm{enc}(\theta)\in\mathbb{R}^{30}$ encode a configuration (ordinal indices for
categorical parameters, values for numeric ones). The EPM is an XGBoost gradient-boosted
tree regressor~\cite{chen2016xgboost} $\hat r=F([\mathbf{x}\,\|\,\mathrm{enc}(\theta)])$,
$F:\mathbb{R}^{50}\to\mathbb{R}$, trained to predict $\log\mathrm{sp}$. It assigns a score
to each (instance, configuration) pair and can therefore rank any candidate set without
solving. Training rows come from the offline sampling of Step~2, and
we deliberately \emph{retain configurations slower than the default} as negative targets:
a model fitted on winners alone cannot recognize a hazard. Online we score all of $\Pi$
and forward the top $k$ by predicted log-speedup, indexed by EPM rank and carrying
portfolio identifiers. The width $k$ trades reachable ceiling against prompt size; we
sweep it in Section~\ref{sec:iteration} and use $k=10$ throughout.

\subsection{Step 4: Foundation-Model Adjudication without Weight Updates}
\label{sec:adjudicate}
The last step is a single foundation-model call that returns an index into the shortlist,
and it is the only component that reads the compressed textual matrix rendering rather
than aggregate features.

Let $\mathbf{x}$ denote the fixed $20$-feature extractor used in this work. If
$\mathbf{x}(I)=\mathbf{x}(I')$, a deterministic selector that observes only
$\mathbf{x}$ must return the same configuration for both instances. When the two
instances prefer different configurations within the portfolio, that selector cannot
recover both instance-specific choices. This limitation is specific to the chosen
representation; alternative finite-dimensional representations may retain the distinction.
PAC-style analyses bound expected regret under a distributional assumption on
instances~\cite{balcan2018learning}. Here we exhibit one concrete
collision under the feature extractor used by our system. The architectural consequence
is that the nominator, which scores from $\mathbf{x}$ alone, returns the \emph{same} $k$
nominees for both members of the pair, while the adjudicator also receives $\psi$ and
therefore has instance-distinguishing information unavailable to the nominator.

\begin{figure}[t]
\centering
\includegraphics[width=\columnwidth]{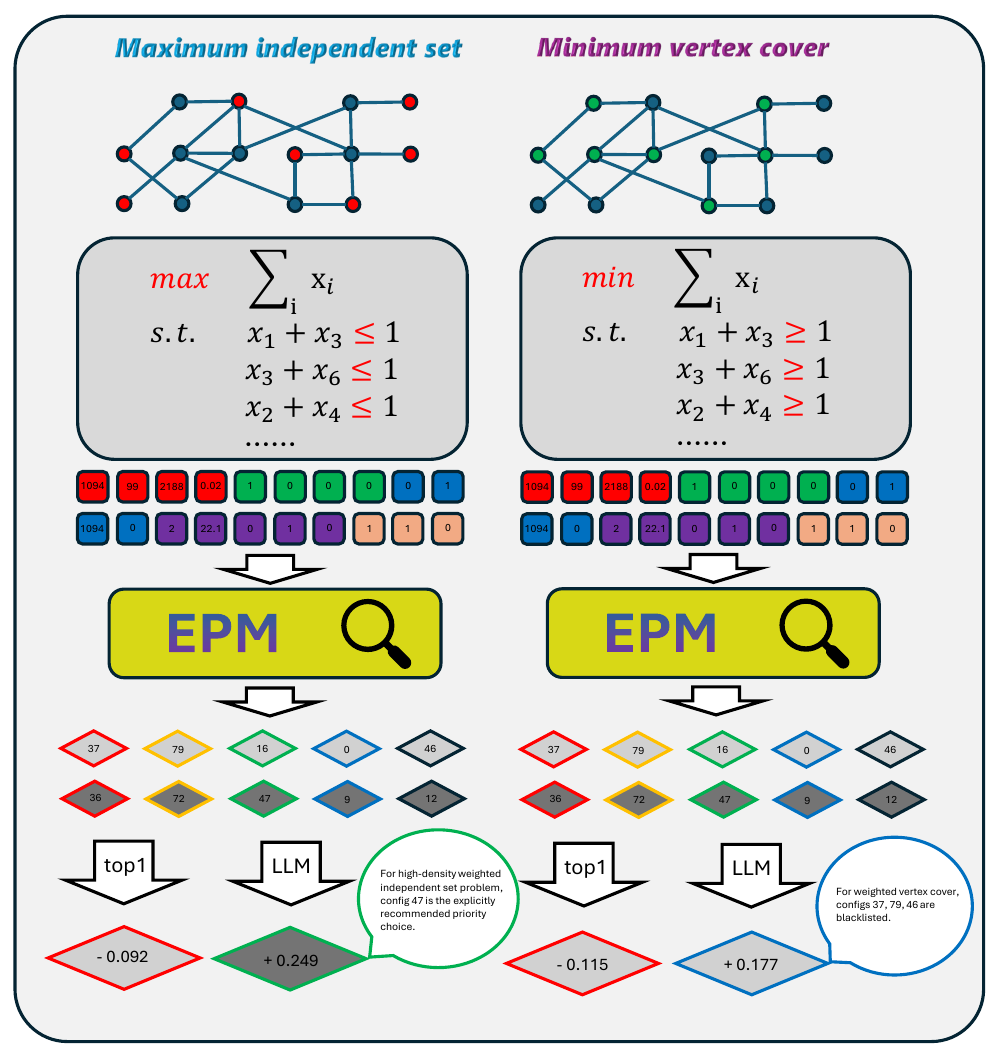}
\caption{Feature degeneracy, illustrated on one shared seed (drawn small; $99$ vertices,
$1094$ edges). The independent-set and vertex-cover models differ only
in constraint sense and objective direction, neither of which reaches the feature vector,
so their $20$ features are byte-identical and the nominator must return one shortlist for
both. Their measured preferred portfolio configurations differ, and reading the matrix
lets the adjudicator pick a different configuration for each (here identifiers
$47$ and $0$, against a shared top nominee $37$ that is slower than the default on both).
The numbers illustrate one pair and provide no aggregate estimate.}
\label{fig:isvc}
\end{figure}

\emph{A pair that exposes the collision.}
\label{sec:isvc}
A generated pair exhibits this collision. From one shared seed our generators build a
maximum-weight independent-set instance and a minimum-weight vertex-cover instance on the
\emph{same} graph. They differ only in constraint sense ($x_i+x_j\le 1$ against
$x_i+x_j\ge 1$) and objective direction, and neither difference reaches the feature
vector, so the two feature vectors come out byte-identical and the pair is
feature-degenerate. The two classes do prefer different configurations, along a
mechanistically sensible axis: independent set is a packing polytope on which selecting
two adjacent vertices is an immediate conflict, so global conflict analysis pays off.
The vertex-cover instance shows no corresponding benefit. Fig.~\ref{fig:isvc} shows one such pair.

This pair exposes information absent from the aggregate numbers. The adjudicator reads the
constraint matrix $\psi$, a channel the features discard, and is conditioned on a
textual memory carried in its prompt. With $\psi$ removed, the two inputs coincide in the
features, shortlist and shared memory, so the prompt contains no instance-specific signal
that favors different choices. Supplying $\psi$ makes such an instance-conditioned choice
possible under our prompt design. Alternative feature representations may carry the same
distinguishing information; this observation motivates the
additional structural channel used in this system.

\emph{Inputs and output.} The prompt carries a deterministic, compressed
rendering of the constraint matrix, which supplies $\psi$; the $20$ features;
the default and all $k$ candidates as full $30$-parameter listings labelled by portfolio
identifier; and the skill memory. It returns strict JSON with a chosen index and a
one-line rationale, and a parse failure falls back to the top nominee. Restricting the
model to a $k$-way choice bounds the output space and prevents it from inventing unmeasured
parameter values. This output-control property provides no guarantee that every selected
configuration solves successfully on every instance.

\emph{The matrix rendering $\psi$.} The channel that carries structure is a text
rendering of the constraint matrix, produced deterministically. In distribution,
constraints are grouped by name after stripping trailing indices. Out of distribution, we
use the name-independent structural grouping of Section~\ref{sec:miplib}. Each group is
represented by one row written as an explicit linear expression and annotated with the
group's cardinality and the range of its right-hand sides. The header reports the counts of
binary, integer and continuous variables. The objective is listed in full when it has at
most $40$ nonzero terms and summarized by its coefficient range otherwise.
Very long rows are truncated, but their \emph{sense and bounds are always printed in
full}, so truncation never conflates a $\le$ row with a $\ge$ one. This preserves the
distinction on which the collision in Section~\ref{sec:principle} turns. On the generated families it occupies a
few hundred tokens per instance.

\emph{Two-objective protocol.} The prompt states an ordered objective: never choose a
candidate at plausible risk of being slower than the default, and deviate from
$\theta^{(1)}$ only with concrete experience-backed evidence, choosing $\theta^{(1)}$
when unsure. Choosing $\theta^{(1)}$ incurs no extra solve because it is the solve the
nominator would have made. The protocol therefore only spends deviations it can justify.

\emph{Skill memory: cold start.} The only evolving artifact is a textual prompt fragment;
foundation-model weights remain fixed. One fully solved shortlist batch supplies every
nominee's runtime and hindsight winner. From these trajectories, one model call per family
induces identifier profiles, veto rules, observable-threshold rules and a decision
checklist. Family indexing lets later rounds retrieve and rewrite one section in isolation.

\begin{figure*}[t]
\centering
\includegraphics[width=\textwidth]{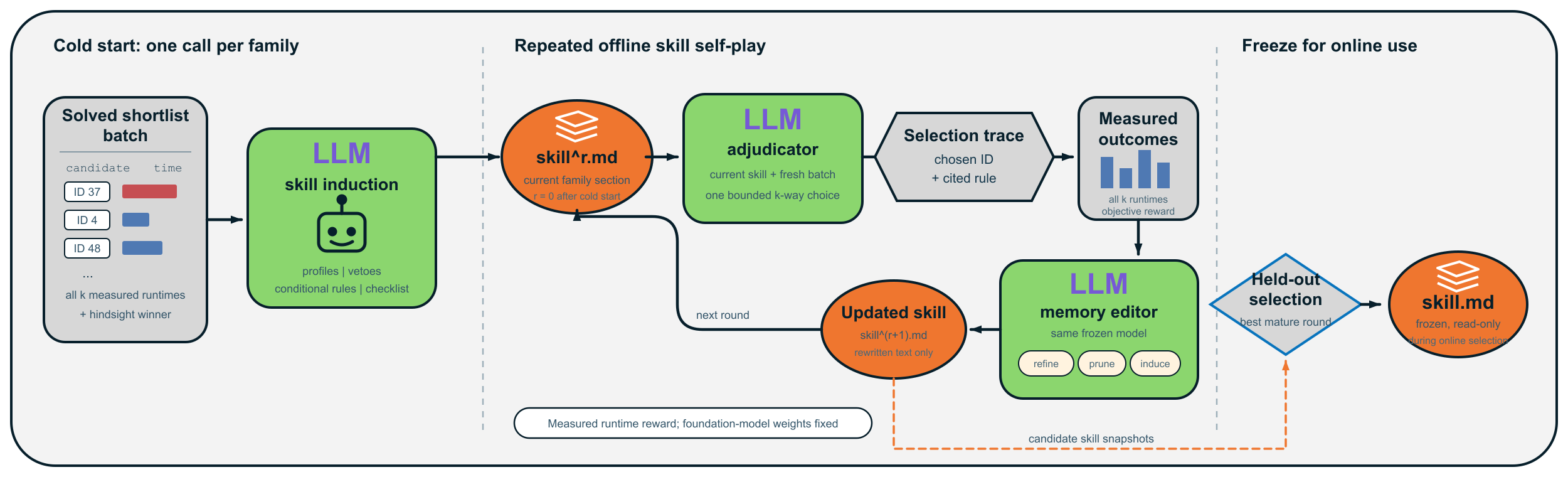}
\caption{Offline construction and refinement of textual skill memory. Cold start induces
family-indexed rules from a solved shortlist batch. Later rounds refine, prune and induce
rules from measured nominee outcomes; a held-out batch selects the snapshot used online.
Model weights remain fixed.}
\label{fig:skillmemory}
\end{figure*}

\emph{Skill memory: the self-play loop.} In each later round, the fixed model adjudicates a
fresh batch, then receives its choice, cited rule and every nominee's measured outcome. It
rewrites the relevant family section by refining rules behind losing deviations, pruning
rules that stopped paying, and inducing rules for uncovered structure. The measured-runtime
reward makes each edit verifiable; only the prompt changes. Figure~\ref{fig:skillmemory}
summarizes the cold-start, refinement and freeze stages.

\section{Experiments}
\label{sec:experiments}

\subsection{Setup}
\label{sec:setup}
Two separate components share the protocol below: the skill iteration and ablations of
Sections~\ref{sec:iteration}--\ref{sec:ablations}, on generators for $16$ self-built
problem families, and the baseline comparison of Section~\ref{sec:baseline}, on the
released Ecole suite. Instances are drawn so the default solve time falls in a fixed
$0.5$--$5$s window, which keeps speedups measurable and prevents timeouts from
dominating. Evaluation batches are seed-disjoint from the pool-building and
EPM-training batches. Solves are serial with single-thread pinning, a fixed solver
seed and a $10$s wall-clock limit. The adjudicator is a Qwen-Max chat model reached
through an OpenAI-compatible API at temperature $0.1$ with an output budget of $4096$
tokens; each instance is decided by exactly one call, and the same model and settings
are used for every reported number unless stated otherwise. We report every head-to-head
comparison under a \emph{practical-difference filter} that declares a pair tied whenever
the two runtimes differ by less than $20\%$. We use this threshold operationally and draw
no inference about measurement variance or statistical significance from it. A configuration
that makes the solver itself abort receives the timeout score. The single occurrence was
on a never-selected member in the width sweep.

\subsection{Skill Iteration: Select the best $k$.}
\label{sec:iteration}
\begin{figure}[!t]
\centering
\includegraphics[width=\columnwidth]{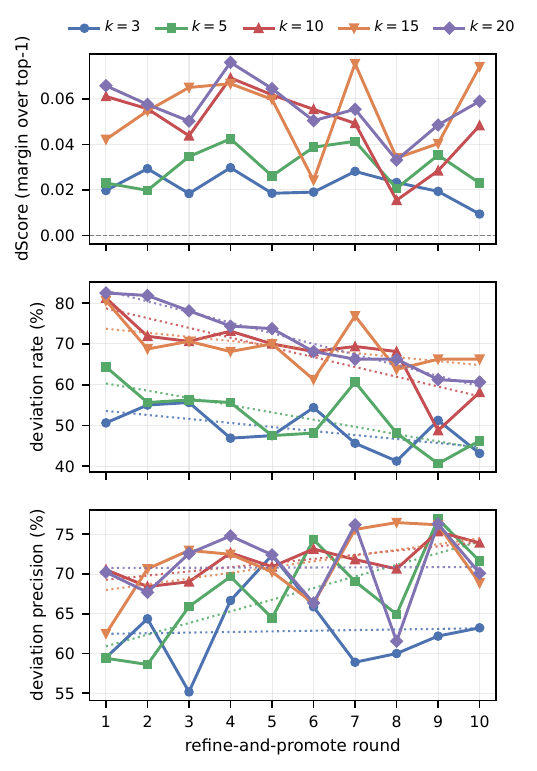}
\caption{Per-round record of all five width runs over ten rounds. Top: margin over the
top nominee. Middle: deviation rate (dotted linear trends). Bottom: deviation precision.
Curves are ordered by $k$.}
\label{fig:iteration}
\end{figure}

We ran the loop autonomously at five shortlist widths, $k\in\{3,5,10,15,20\}$, each for
ten refinement rounds on disjoint $160$-instance batches. Each round
evaluates the current skill on a fresh batch, refines from \emph{that batch alone},
and advances.
Fig.~\ref{fig:iteration} reports the per-round record of all five runs, and
Table~\ref{tab:ksweep} compares the widths on the batches all five share.

\begin{figure}[!ht]
\centering
\includegraphics[width=\columnwidth]{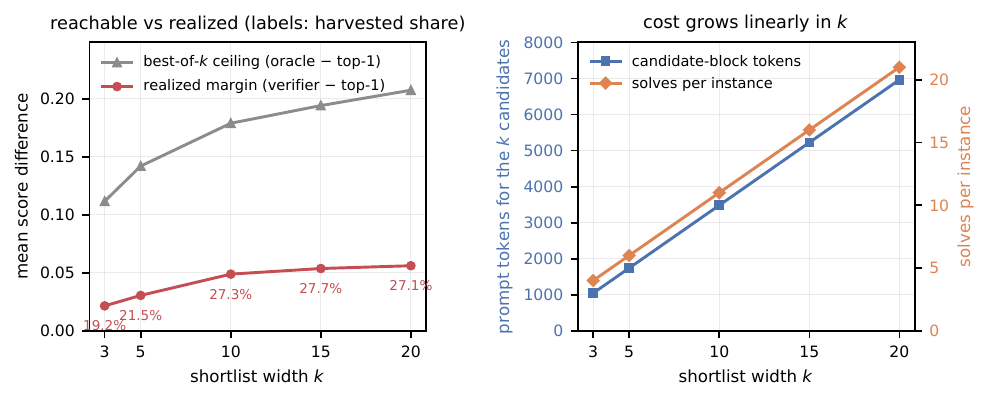}
\caption{Why $k=10$. Left: widening the shortlist raises the best-of-$k$ ceiling
steadily, but the realized margin flattens after $k=10$; labels give the
share of the ceiling harvested, which sits at about $27\%$ for every $k\geq10$. Right:
both costs (prompt tokens and solves per batch) are linear in $k$. Past $k=10$ the
added ceiling yields negligible realized speedup.}
\label{fig:ksweep}
\end{figure}

\begin{figure}[!ht]
\centering
\includegraphics[width=\columnwidth]{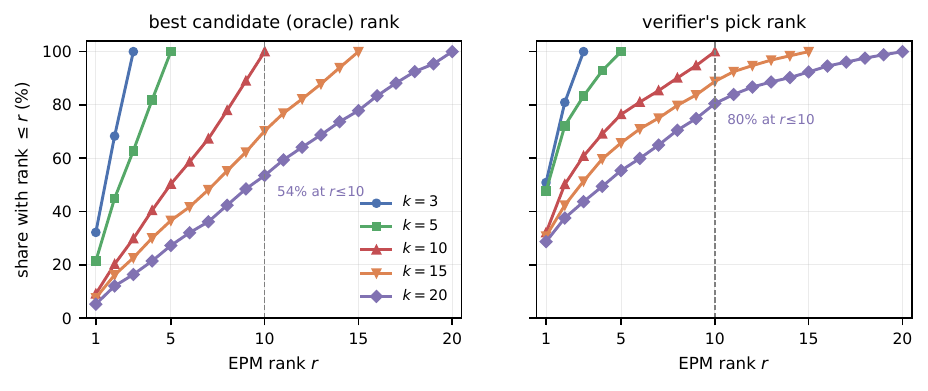}
\caption{Where the shortlist mass sits, on the batches shared by all five runs. Left:
the rank of the best candidate; right: the rank the adjudicator picks. At $k=20$ the best
candidate often lies past rank $10$ and the adjudicator does reach into that tail, but those
extra candidates add no margin (Table~\ref{tab:ksweep}).}
\label{fig:oraclerank}
\end{figure}

\emph{Why $k=10$.} Raising $k$ expands the best-of-$k$ headroom but also increases
solve and prompt cost. In a controlled sweep over $k\in\{3,5,10,15,20\}$, the portfolio,
EPM training data and evaluation seeds are shared; only the skill document names $k$.
The harvested fraction plateaus near $27\%$ at $k=10$, $15$ and $20$ ($27.3\%$, $27.7\%$
and $27.1\%$; Fig.~\ref{fig:ksweep}). Although the $k=20$ adjudicator reaches beyond rank
$10$ (Fig.~\ref{fig:oraclerank}), the additional candidates are not separated by the
available evidence, and doubling the width from $10$ to $20$ adds only $+0.007$ margin.
We therefore use $k=10$ in Section~\ref{sec:baseline}, the smallest evaluated width at
which the harvested fraction saturates and the shortlist remains cheaper to solve and
describe.

\emph{Conservative refinement produces a mid-run peak.} Across all five runs, the
deviation rate falls from the first five rounds to the last, while deviation precision
rises or holds (at $k=10$, $73.4\%\!\to\!62.5\%$ and
$70.3\%\!\to\!73.0\%$, respectively; Fig.~\ref{fig:iteration}, middle and bottom).
Losing deviations provide evidence to prune unprofitable rules, so later memories
deviate less without becoming less precise. The margin peaks around rounds three to five,
when deviations remain frequent but are already precise, then eases as refinement favors
caution.

\emph{Nonmonotone refinement.} When a family is
repaired by vetoing one hazardous identifier, the adjudicator can shift to a
\emph{different} hazardous identifier in the same family later. At $k=10$ the residual
damage concentrates in one family: cutting stock swings from $+0.091$ in round two to
$-0.177$ in round six as timeouts migrate between configurations. This is a
\emph{coverage-limited} family, where several members can explode and per-identifier
vetoes play whack-a-mole. Per-identifier refinement therefore leaves the family-level
risk unresolved. We freeze the skill at the best mature round, chosen on a
held-out batch.

\begin{table}[!t]
\centering
\caption{Shortlist width sweep on the ten $160$-instance batches completed in all
five runs ($1600$ instances each). Scores are means over instances; \emph{margin} is
adjudicator minus top-1, \emph{headroom} is oracle minus top-1, and \emph{harvested}
is margin as a share of headroom. Best width per row in bold where
one exists; harvested is unbolded because it is flat from $k=10$ on.}
\label{tab:ksweep}
\small
\setlength{\tabcolsep}{3pt}
\begin{tabular}{@{}lccccc@{}}
\toprule
& $k=3$ & $k=5$ & $k=10$ & $k=15$ & $k=20$ \\
\midrule
mean score & 0.401 & 0.409 & 0.426 & 0.431 & $\mathbf{0.434}$ \\
top-1 & 0.380 & 0.379 & 0.377 & $\mathbf{0.378}$ & $\mathbf{0.378}$ \\
margin & $+0.021$ & $+0.031$ & $+0.049$ & $+0.054$ & $\mathbf{+0.056}$ \\
oracle & 0.491 & 0.521 & 0.556 & 0.572 & $\mathbf{0.585}$ \\
headroom & 0.112 & 0.142 & 0.179 & 0.194 & $\mathbf{0.207}$ \\
harvested & $19.2\%$ & $21.5\%$ & $27.3\%$ & $\mathbf{27.7\%}$ & $27.1\%$ \\
win rate & $87.9\%$ & $87.9\%$ & $90.2\%$ & $90.4\%$ & $\mathbf{90.7\%}$ \\
dev.\ rate & $49.1\%$ & $52.3\%$ & $67.9\%$ & $69.2\%$ & $71.3\%$ \\
dev.\ prec. & $62.8\%$ & $67.0\%$ & $\mathbf{71.5\%}$ & $71.2\%$ & $70.8\%$ \\
\bottomrule
\end{tabular}
\end{table}

Deviation rate and precision require within-$k$ interpretation. Widening the
shortlist mechanically lowers the chance the top nominee is the true best, so a larger
$k$ raises the deviation rate even a perfectly calibrated selector would show (the
$49\%\!\to\!71\%$ rise in Table~\ref{tab:ksweep}). Rates here are therefore read within a
fixed $k$, never across.

\subsection{Ablations}
\label{sec:ablations}

\paragraph{The shortlist leaves substantial selection headroom}
\label{sec:coverage}
Two measurements characterize the sampled portfolio and its ordering. First, the nominee
sets are strongly family-structured
(mean Jaccard overlap $0.518$ within a family against $0.090$ across), each family
spreading over $3$ to $16$ identifiers, with a few all-rounders beneath family-specific
members. This shows that the realized pool supplies different candidate sets across the
evaluated generator families. Second, an oracle over the same $10$ nominees scores $0.556$
against the top nominee's $0.377$, and its position in the shortlist is almost flat
(Table~\ref{tab:ablation}). These observations support selection headroom within the
sampled pool, while near-optimal portfolio coverage remains untested.

\paragraph{What the LLM-based adjudicator adds under this protocol}
\label{sec:main}
Table~\ref{tab:ablation} holds the $10$ nominees, instances and measured solve outcomes
fixed while comparing chooser implementations. The with-LLM arm is the complete
adjudicator bundle: matrix rendering, textual skill memory and the foundation-model call;
the no-LLM arm returns the EPM's top nominee. The comparison therefore measures the
bundle's incremental value over feature-only top-1 selection; individual causal
contributions from the matrix, memory and model remain unidentified. Under this protocol, no-LLM is close to random
choice within the shortlist, whereas the complete adjudicator attains a higher score by
beating the top nominee more often than it loses across all ten batches while harvesting
only part of the oracle headroom. The observed deviations also convert more
default-losing instances into wins than the reverse. Every option remains a member of the
fixed, previously evaluated portfolio, so the model chooses among existing configurations
and generates no parameter values. Instance-level timeout or failure risk remains.

\begin{table}[!t]
\centering
\caption{The ablation: chooser implementations over a fixed shortlist. All arms use the
same $10$ nominees on the same $1600$ held-out instances. The with-LLM arm comprises the
matrix, skill memory and foundation-model adjudicator, while no-LLM returns the EPM top
nominee; a hindsight oracle over the same ten scores $0.556$. Lower block: with-LLM
against no-LLM, per instance.}
\label{tab:ablation}
\small
\begin{tabular}{lccc}
\toprule
 & Random & no-LLM & \textbf{with-LLM} \\
\midrule
Mean score & $0.365$ & $0.377$ & $\mathbf{0.426}$ \\
Win rate vs.\ default & $84.9\%$ & $84.9\%$ & $\mathbf{90.2\%}$ \\
Median speedup & $2.34\times$ & $2.43\times$ & $\mathbf{2.69\times}$ \\
SGM speedup & $1.35\times$ & $1.37\times$ & $\mathbf{1.42\times}$ \\
Default time saved & $41.1\%$ & $43.8\%$ & $\mathbf{50.4\%}$ \\
Picks the best of $10$ & $10.0\%$ & $9.1\%$ & $\mathbf{22.5\%}$ \\
Headroom harvested & $-7\%$ & $0\%$ & $\mathbf{27\%}$ \\
\midrule
\multicolumn{4}{l}{\emph{with-LLM against no-LLM, per instance}} \\
Faster : slower & \multicolumn{3}{c}{$\mathbf{771 : 316}$} \\
\quad above threshold ($>\!20\%$) & \multicolumn{3}{c}{$\mathbf{538 : 185}$} \\
Rescues : breaks & \multicolumn{3}{c}{$\mathbf{121 : 37}$} \\
Batches favouring with-LLM & \multicolumn{3}{c}{$\mathbf{10/10}$} \\
\bottomrule
\end{tabular}
\end{table}

\begin{figure*}[!t]
\centering
\includegraphics[width=0.98\textwidth]{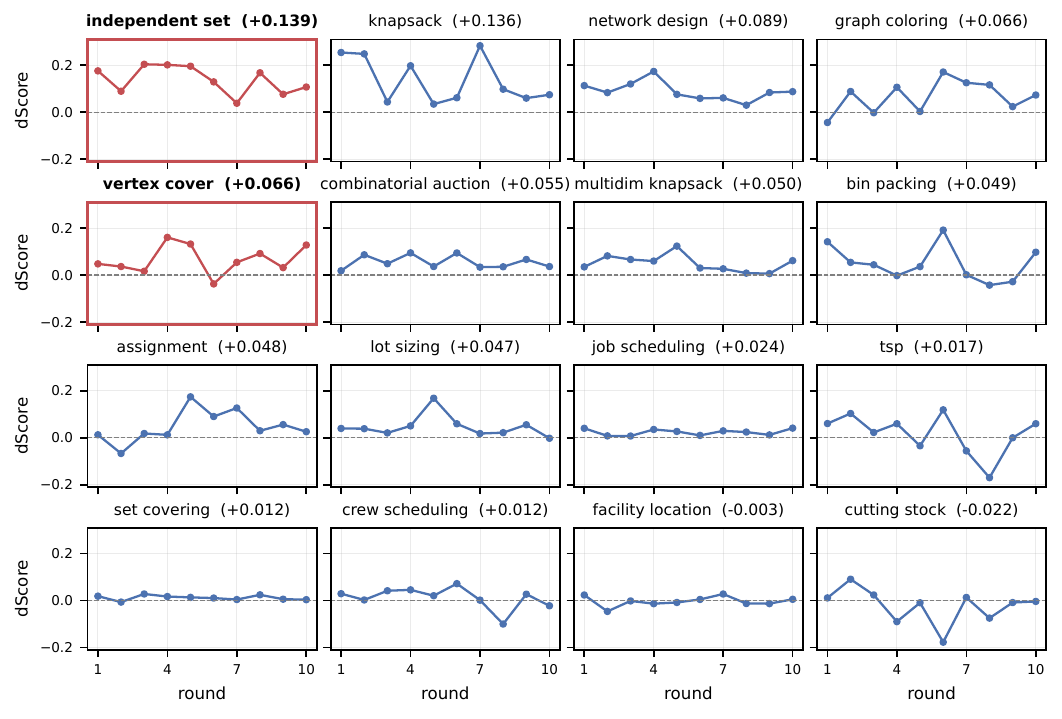}
\caption{Margin over the top nominee by family, round by round, at the selected
width $k=10$; panels are ordered by mean margin, which is given in each title, and
share one vertical scale. Fourteen of sixteen families are positive on average and eight
never dip below zero; the damage concentrates in cutting stock.}
\label{fig:family}
\end{figure*}

\paragraph{Where the gain lives}
\label{sec:family}
The aggregate varies by family (Fig.~\ref{fig:family}): fourteen of sixteen
families are positive on average and eight in \emph{every} round. The strongest and
steadiest are independent set, knapsack and network design. Note where the degeneracy
pair of Section~\ref{sec:isvc} lands. That construction establishes a collision under the
fixed $20$-feature representation for one IS/VC pair. At family level, independent set is
the strongest of all sixteen families and vertex cover is positive in nine of ten rounds.
This alignment is consistent with the hypothesis that the additional matrix channel can
help when aggregate features omit relevant structure. The one-pair construction leaves
collision prevalence and a family-wide causal mechanism unmeasured. The only two negative families,
facility location and cutting stock, are
the coverage-limited case of Section~\ref{sec:iteration}, where several portfolio members
can time out and vetoing one merely moves the hazard. Between the extremes lie saturated
families such as set covering and job scheduling, whose nominees are nearly
interchangeable, leaving little room for a selector over the evaluated shortlist to
improve on the top nominee.

\subsection{Baseline Comparison}
\label{sec:baseline}

\begin{table*}[t]
\centering
\caption{Baselines under a matched one-pass protocol, $400$ held-out instances, all
arms re-measured in one session against a shared default (round-robin, median of three).
Per row, \rkone{gold} is first, \rktwo{silver} second, \rkthree{bronze} third. $\Delta$
is our mean score minus the arm's; ``above threshold'' counts instances whose runtimes differ by
more than the $20\%$ practical-difference threshold. \emph{no-LLM} returns the EPM's top nominee from the same
shortlist and portfolio; the matrix--memory--foundation-model adjudicator bundle is absent.}
\label{tab:baseline}
\begin{tabularx}{\textwidth}{l CCCCCC}
\toprule
 & default & ours ($k=10$) & no-LLM & ISAC+ & ISAC+SMAC & ISAC+ROAR \\
\midrule
\multicolumn{7}{l}{\emph{Aggregate, all four families ($n=400$)}}\\
Mean score              & $0.0000$ & \rkone{$0.5638$} & \rkthree{$0.5429$} & \rktwo{$0.5441$} & $0.4933$ & $0.4318$ \\
SGM speedup             & $1.00$   & \rkone{$4.48$}   & \rkthree{$4.28$}   & \rktwo{$4.29$}   & $3.66$   & $3.04$ \\
Win rate vs.\ default   & ---      & \rkone{$99.0\%$} & \rkthree{$97.8\%$} & \rktwo{$98.0\%$} & $96.8\%$ & $95.5\%$ \\
Default time removed    & ---      & \rkone{$69.6\%$} & \rktwo{$66.4\%$}   & \rkthree{$66.0\%$} & $62.8\%$ & $59.2\%$ \\
$\Delta$ vs.\ ours      & ---      & ---              & $+0.0209$          & $+0.0197$        & $+0.0705$ & $+0.1321$ \\
Above threshold ($>\!20\%$), W:L & --- & ---           & $92\!:\!58$        & $83\!:\!52$      & $163\!:\!52$ & $191\!:\!39$ \\
Solver-call budgets & --- & 6000 & 6000 & 6000 & 6000 & 6000 \\
\midrule
\multicolumn{7}{l}{\emph{Mean score by family ($n=100$ each)}}\\
facility location       & $0.0000$ & \rkone{$0.6160$} & \rktwo{$0.5985$}   & $0.5262$         & $0.5462$         & \rkthree{$0.5839$} \\
independent set         & $0.0000$ & \rkone{$0.3703$} & \rkthree{$0.3395$} & \rktwo{$0.3620$} & $0.3358$         & $0.3391$ \\
set covering            & $0.0000$ & \rktwo{$0.7048$} & \rkthree{$0.6942$} & \rkone{$0.7105$} & $0.6287$         & $0.5521$ \\
combinatorial auction   & $0.0000$ & \rktwo{$0.5641$} & \rkthree{$0.5393$} & \rkone{$0.5777$} & $0.4627$         & $0.2520$ \\
\bottomrule
\end{tabularx}
\end{table*}

This self-contained comparison is separate from
Sections~\ref{sec:iteration}--\ref{sec:ablations}. It runs on the four
standard Ecole generators: set covering, combinatorial auction, capacitated facility
location and maximum independent set~\cite{gasse2019exact}, from the released ML-for-MILP
benchmark. It uses its own offline artifacts and its own $400$ held-out instances. The
sixteen self-built families and their skill memory play no part here. The two components
share only the instance-independent portfolio $\Pi$ and the protocol of
Section~\ref{sec:setup}. Their numbers are therefore reported separately and never read
across.

We compare against the default and three points on the ISAC family, organized as a
descriptive, non-factorial ladder. The ladder compares complete systems and carries no
component-level causal attribution. All arms are one-pass
(features in, one configuration out, no solver call to choose) and receive the same
offline solve budget. \emph{ISAC+ROAR}~\cite{kadioglu2010isac} clusters instances in our
feature space and tunes one configuration per cluster by model-free
racing~\cite{hutter2011smac}, realizing a piecewise-constant map; we sweep the cluster
count $c$ and select it on held-out \emph{training} instances, never on test.
\emph{ISAC+SMAC} substitutes model-based SMAC~\cite{hutter2011smac} inside each cluster,
which strengthens the within-cluster search while leaving the one-configuration-per-cluster
constraint intact. The change from ISAC+ROAR to ISAC+SMAC reports how the observed margin
moves under stronger within-cluster search.
\emph{ISAC+}~\cite{ansotegui2014isacplus} pools those configurations into a portfolio and
selects per instance, discarding that constraint; we further give it our feature set and
surrogate model class and let it re-select $c$ on validation. Matching these components
makes ISAC+ our closest baseline, but the systems still differ in portfolio construction
and selector design; their gap is therefore an end-to-end comparison.

The baseline ladder progressively strengthens the comparison
(Table~\ref{tab:baseline}). Replacing ROAR with SMAC inside each cluster halves the gap;
allowing per-instance selection in ISAC+ closes most of what remains. Ours leads every
aggregate metric, but against ISAC+, the best-performing evaluated per-instance baseline
under the matched protocol, the mean-score margin is $+0.020$. This is similar to the
$+0.021$ gain of the complete adjudicator over no-LLM.

The bounded score compresses this residual because it saturates near the top and averages
families with little shortlist headroom. In uncompressed aggregates, the ISAC+ gap is
$+0.19$ in SGM speedup ($4.48$ vs.\ $4.29$) and $+3.6$ points of default wall-clock
removed. Above the $20\%$ practical-difference threshold, the descriptive win--loss count
is $83\!:\!52$; we make no statistical-significance inference from it.

Family outcomes track shortlist headroom only partly. Set covering is saturated
(headroom $0.031$), whereas combinatorial auction retains headroom ($0.091$) but converts
little of it; the data cannot attribute this shortfall among shortlist composition, EPM
ordering and adjudication. Facility location and independent set have the largest
headroom and disagreement with ISAC+, yet $>\!20\%$ comparisons favor us only on the
former. Runtime dispersion and headroom are therefore incomplete proxies for
discriminability. The facility-location result is consistent with matrix-visible
structure aiding selection, but this is not the only possible explanation; on independent
set, disagreements cancel and configuration yields the least benefit.

The cluster-level arms expose the expected granularity limit. At $c=4$, ISAC+ROAR assigns
$247$ instances, including every independent-set instance and $95$ of $100$
combinatorial-auction instances, to one configuration. Its combinatorial-auction score is
$0.252$ against our $0.564$; splitting more finely at $c=8$ raises ISAC+SMAC to $0.463$.

\subsection{Transfer to Two MIPLIB Instances}
\label{sec:miplib}
The experiments above use a closed set of generators. To probe bounded transfer without a
statistical generalization claim, we retain two heterogeneous, unlabeled MIPLIB
2017 instances~\cite{gleixner2021miplib} whose $20$ features all lie within the EPM
training ranges. This outcome-independent, pre-solve criterion avoids confounding transfer
with EPM extrapolation. The portfolio and skill memory come from the synthetic families;
nothing is refit on MIPLIB, and cluster-level baselines are likewise trained on the
generators and applied unchanged. All arms therefore face the same shift in one pass.
Because both instances remain open at short time scales, we compare time to an attainable
relative gap. Targets are instance-specific: \texttt{mas76} reaches $0.1\%$ in minutes,
whereas \texttt{50v-10} cannot get below a few percent, making a common target trivial for
one or unreachable for the other.

\begin{table}[t]
\centering
\caption{Wall-clock time (s) to an instance-specific relative gap on two
feature-range-compatible MIPLIB 2017 cases; lower is better. Each instance's five arms
share one measurement session and default run.}
\label{tab:miplib}
\small
\begin{tabularx}{\columnwidth}{l CC}
\toprule
time to gap (s) & \texttt{mas76} & \texttt{50v-10} \\
                & (gap $0.001$)  & (gap $0.15$)   \\
\midrule
SCIP default              & $172.0$          & $68.0$ \\
ISAC+ROAR                 & $123.2$          & $83.0$ \\
ISAC+SMAC                 & $363.8$          & $55.0$ \\
ISAC+                     & $428.6$          & $38.0$ \\
\textbf{ours}             & $\mathbf{87.2}$  & $\mathbf{22.0}$ \\
\midrule
ours speedup vs.\ default & $\mathbf{1.97\times}$ & $\mathbf{3.09\times}$ \\
\bottomrule
\end{tabularx}
\end{table}

On both instances, ours reaches the target faster than the default and all three
competitive ISAC-based methods under the same transfer protocol, by about $2\times$ on
\texttt{mas76} and $3\times$ on \texttt{50v-10} (Table~\ref{tab:miplib}). Each rationale
is traceable to the memory. For \texttt{mas76}, the adjudicator maps a dense binary,
medium-CV matrix with at least ten rows to multidimensional knapsack, then selects $48$
and vetoes $33$, both literal rules of that class. For \texttt{50v-10}, the renderer
identifies lot sizing and network design from flow conservation and capacity/big-M
inequalities; the adjudicator selects $4$, a core lot-sizing entry, and rejects $0$ under
a network-design veto. Each decision thus links observed structure to a recorded class
and rule.

These cases support only a narrow conclusion: experience distilled from synthetic classes
can transfer to real, unlabeled models through structural reading rather than label
matching, as required by the open-set argument of Section~\ref{sec:principle}.

\section{Discussion and Conclusion}
\label{sec:discussion}

The evidence supports a mechanistic interpretation within the evaluated setting: reusable
candidate construction followed by fine ranking among structurally subtle candidates. The
sampled pool and EPM shortlist provide useful headroom on the evaluated families, while
broader portfolio coverage remains open. Three remarks define the scope.

\emph{Big Data and latency scope.} We operationalize Variety through heterogeneous MIP
structure and Value through solver performance. The one-call, zero-solve selection
interface addresses Velocity, while volume scaling, distributed storage, streaming
ingestion and end-to-end latency remain unmeasured. Deployment accounting must include
model-call, serialization and solver-runtime costs.

\emph{Alternative matrix readers remain open.} A bipartite graph neural network ranker over
the full coefficient matrix~\cite{gasse2019exact,cappart2023combinatorial} could in principle
exploit a related structural channel. The current evidence supports improvement over the
specified feature-only top-1 selector. The necessity or uniqueness of a foundation model
remains open; a fair comparison requires a dimension-agnostic architecture and matched
supervision and training budgets.

\emph{Architectural scope.} What we propose is an
architecture that nominates from a fixed random portfolio and then adjudicates with a foundation
model that reads structure absent from the aggregate features. The same stages can
accommodate richer feature maps, parameter spaces and other solvers, and can extend to
settings where a cheap model nominates from a measured portfolio before semantic
tie-breaking. In the matched study, the complete bundle improves over feature-only top-1
selection and leads the best-performing evaluated per-instance baseline
(Tables~\ref{tab:ablation} and~\ref{tab:baseline}).

\emph{Overall.} We factorized per-instance MIP configuration into a randomized nomination
stage from a fixed portfolio and semantic adjudication with matrix and memory inputs. The
realized pool and shortlist provide oracle headroom on the evaluated families, while
near-universal coverage remains untested. An IS/VC pair with byte-identical features but
different preferred configurations exhibits a collision that a deterministic selector
under our $20$-feature map cannot resolve. The complete adjudicator makes a bounded $k$-way
choice in one frozen-model call, with no online solve or gradient update. Against the top
nominee, it records $538$ wins and $185$ losses above the $20\%$ practical-difference
threshold and leads evaluated per-instance baselines under a matched protocol. Whether
refined textual memory transfers to unanticipated structure remains open.

\end{document}